\documentclass[letterpaper]{article} 
\usepackage{aaai2026}  
\usepackage{times}  
\usepackage{helvet}  
\usepackage{courier}  
\usepackage[hyphens]{url}  
\usepackage{graphicx} 
\usepackage{natbib}  
\usepackage{caption} 
\usepackage{algorithm}
\usepackage{algorithmic}

\usepackage{threeparttable}
\usepackage{multirow}
\usepackage{booktabs}
\usepackage{makecell}
\usepackage[table]{xcolor}
\usepackage{pifont}
\usepackage{array}
\usepackage{xcolor}
\usepackage{amsmath}
\usepackage{amssymb}

\usepackage{newfloat}
\usepackage{listings}
\DeclareCaptionStyle{ruled}{labelfont=normalfont,labelsep=colon,strut=off} 
\floatstyle{ruled}
\newfloat{listing}{tb}{lst}{}
\floatname{listing}{Listing}
\title{MUSE: Multi-Scale Dense Self-Distillation for \\ Nucleus Detection and Classification}
\author{
    Zijiang Yang\textsuperscript{\rm 1,2,}\equalcontrib\thanks{This work was done when Zijiang Yang conducted an internship at DAMO Academy, Alibaba Group.},
    Hanqing Chao\textsuperscript{\rm 2,3,5,}\equalcontrib,
    Bokai Zhao\textsuperscript{\rm 2,}\equalcontrib,
    Yelin Yang\textsuperscript{\rm 4},
    Yunshuo Zhang\textsuperscript{\rm 4}, \\
    Dongmei Fu\textsuperscript{\rm 1},
    Junping Zhang\textsuperscript{\rm 5},
    Le Lu\textsuperscript{\rm 2},
    Ke Yan\textsuperscript{\rm 2,3},
    Dakai Jin\textsuperscript{\rm 2},
    Minfeng Xu\textsuperscript{\rm 2},
    Yun Bian\textsuperscript{\rm 4},
    Hui Jiang\textsuperscript{\rm 4}
}
\affiliations{
    \textsuperscript{\rm 1}University of Science and Technology Beijing\\
    \textsuperscript{\rm 2}DAMO Academy, Alibaba Group\\
    \textsuperscript{\rm 3}Hupan Lab\\
    \textsuperscript{\rm 4}Shanghai Institution of Pancreatic Disease\\
    \textsuperscript{\rm 5}Fudan University\\
    \{hanqing.chq, eric.xmf\}@alibaba-inc.com,
    bianyun2012@foxmail.com, jianghui5131@163.com
}

\usepackage{bibentry}

\begin{document}

\maketitle

\begin{abstract}
Nucleus detection and classification (NDC) in histopathology analysis is a fundamental task that underpins a wide range of high-level pathology applications. 
However, existing methods heavily rely on labor-intensive nucleus-level annotations and struggle to fully exploit large-scale unlabeled data for learning discriminative nucleus representations.
In this work, we propose \textbf{MUSE} (\textbf{MU}lti-scale den\textbf{SE} self-distillation), a novel self-supervised learning method tailored for NDC. At its core is \textbf{NuLo} (\textbf{Nu}cleus-based \textbf{Lo}cal self-distillation), a coordinate-guided mechanism that enables flexible local self-distillation based on predicted nucleus positions. By removing the need for strict spatial alignment between augmented views, NuLo allows critical cross-scale alignment, thus unlocking the capacity of models for fine-grained nucleus-level representation. To support MUSE, we design a simple yet effective encoder-decoder architecture and a large field-of-view semi-supervised fine-tuning strategy that together maximize the value of unlabeled pathology images. 
Extensive experiments on three widely used benchmarks demonstrate that MUSE effectively addresses the core challenges of histopathological NDC. The resulting models not only surpass state-of-the-art supervised baselines but also outperform generic pathology foundation models.
\end{abstract}

\begin{links}
    \link{Code}{https://github.com/alibaba-damo-academy/MUSE}
\end{links}

\section{Introduction}

\begin{figure}[t]
    \centering
    \includegraphics[width=0.99\linewidth]{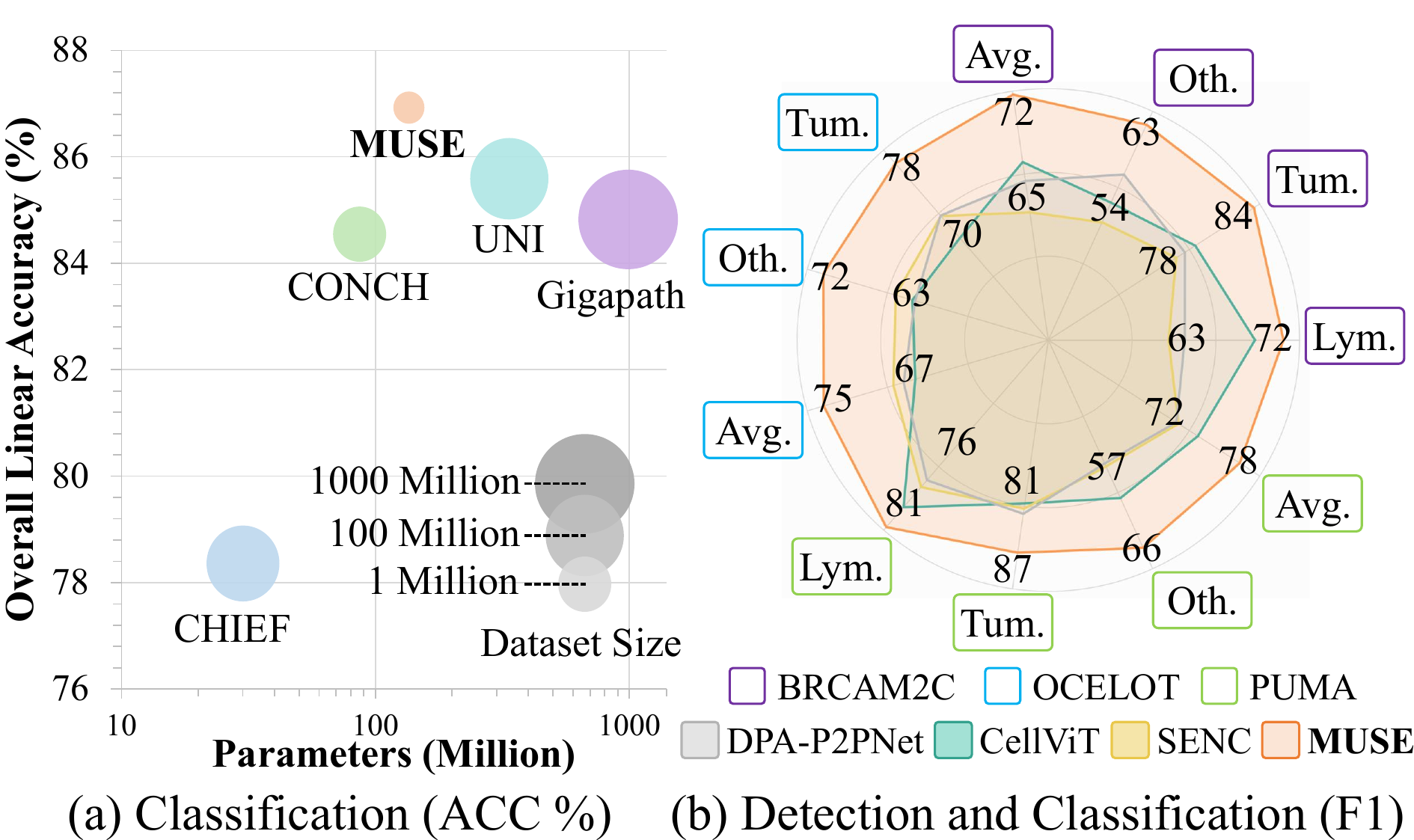}
    \caption{
    Comparison of MUSE and SOTA methods.
    (a) Compared with pathology pretraining methods, MUSE achieves better nucleus classification performance with smaller backbones and only 0.5 million samples. 
    (b) After fine-tuning, MUSE outperforms SOTA methods on nucleus detection and classification.
    Lym., Tum., Oth., and Avg. denote the F1 score of lymphocytes, tumor nucleus, other nucleus, and average, respectively.
    }
    \label{fig:first_fig}
\end{figure}

Nucleus detection and classification (NDC) is a foundational task in histopathological diagnosis~\cite{page2023spatial,zhang2025systematic}. Core pathological workflows, including disease diagnosis, biomarker evaluation, and prognosis prediction, critically depend on the precise localization and identification of specific types of nuclei~\cite{wang2023deep,corredor2019spatial}. In histopathology, accurate recognition of nucleus types relies on both nuclear morphology and the structural context of the surrounding tissue. However, manual annotation for NDC is extremely labor-intensive and time-consuming. To reduce annotation costs while ensuring representativeness, most existing datasets annotate only small high-magnification tiles (typically around 128 $\mu$m per side, containing dozens to hundreds of nuclei). Despite this compromise, these annotated samples remain insufficient to capture the full variability in tissue architecture, nuclear morphology, and staining conditions, limiting the generalizability of the supervised models~\cite{hovernet, cellvit}. Furthermore, the inherently limited field of view (FoV) in such small tiles restricts access to broader tissue-level context, further contributing to performance bottlenecks~\cite{chai2025review,cellvit++}.

\begin{figure}[t]
    \centering
    \includegraphics[width=0.99\linewidth]{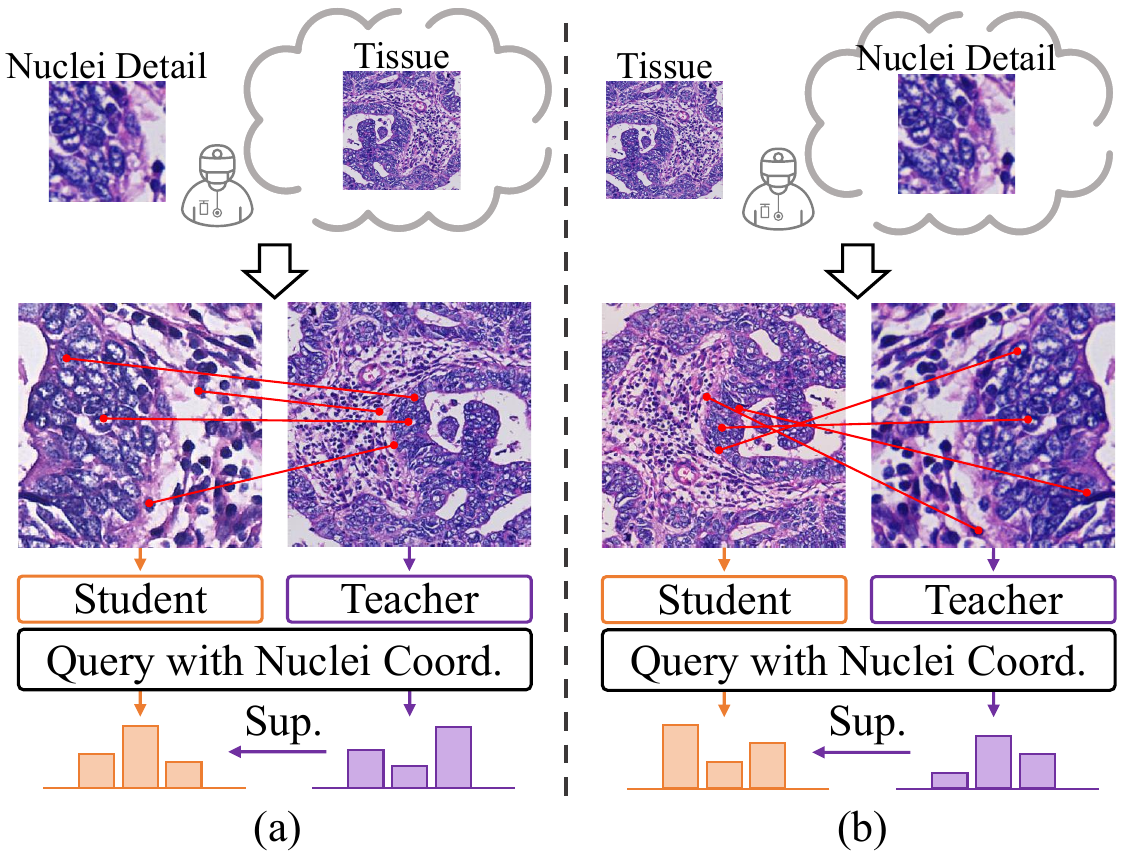}
    \caption{
    Motivation of NuLo.
    Pathologists can flexibly associate nuclei with tissue.
    This inspires us to introduce two cross-scale self-distillation processes on matched nuclei:
    (a) inferring tissue information with nuclei detail and (b) inferring nuclei detail with tissue information.
    }
    \label{fig:motivation}
\end{figure}

To alleviate these limitations, recent studies have explored the integration of unlabeled data. Some approaches incorporate additional low-magnification images with larger FoVs to provide richer contextual information~\cite{dpap2pnet}, while others employ semi-supervised learning techniques such as pseudo-labeling and consistency regularization~\cite{su2023dual,bai2021novel}. However, these methods still depend heavily on the limited set of labeled data. Large-FoV (LFoV) inputs enrich contextual information per tile but do not expand the diversity or quantity of training samples. Meanwhile, semi-supervised learning methods typically assume that labeled and unlabeled data are drawn from the same distribution~\cite{yang2022survey}, which constrains the inclusion of diverse unlabeled samples and often leads to unstable performance when this assumption is violated. Recently, self-supervised learning (SSL) has emerged as a powerful paradigm for learning generalizable visual representations from large-scale unlabeled data~\cite{ibot,dinov2}. In computational pathology, SSL-pretrained foundation models have demonstrated promising improvements across various downstream tasks~\cite{uni,chief}, offering a potential path forward for addressing the dilemma faced by NDC. However, most existing pathology foundation models directly adopt SSL methods originally developed for natural images (e.g., DINOv2~\cite{dinov2}), primarily focusing on image-level representation learning without adapting to the local nucleus-level demands of dense prediction tasks such as NDC. While some of these models are trained on millions~\cite{uni} or even billions~\cite{gigapath} of histology tiles, they often struggle to capture discriminative nucleus-level features, limiting their effectiveness for NDC (Figure~\ref{fig:first_fig}a). This limitation arises from three key issues. First, existing SSL methods typically enhance local representations through patch-level masked image modeling (e.g., iBOT~\cite{ibot}). However, these methods require strict spatial alignment between patch tokens, limiting essential spatial augmentations such as scale jittering, which is crucial for learning local-to-global alignment~\cite{dino}.
Second, current SSL models typically lack supervision across different feature levels. As NDC is a dense prediction task, existing studies have shows that it benefits from multi-level feature fusion~\cite{dpap2pnet}. Third, most foundation models operate on small pathology tiles with a limited FoV (e.g., 256×256 pixels, $\approx$16.4K $\mu$m$^2$), which restricts their capacity to capture broader tissue context.

In this work, we propose MUSE (\textbf{MU}lti-scale den\textbf{SE} self-distillation), a novel SSL method tailored for NDC. Built around this pretraining strategy, we develop a simple yet effective framework encompassing a unified encoder-decoder model architecture, pretraining strategy, and downstream fine-tuning pipeline that collectively enable efficient utilization of both annotated and unannotated data. We first introduce a lightweight and flexible encoder-decoder backbone that supports variable input sizes (from 96 to 1024 pixels), enabling large-FoV training and multi-level feature fusion. Based on this, MUSE incorporates a novel Nucleus-based Local Self-distillation (NuLo) mechanism, which enhances global SSL with flexible local self-distillation. NuLo employs a lightweight nucleus detector to estimate nuclear coordinates and performs local self-distillation based on feature interpolation around each nucleus. This coordinate-guided local self-distillation removes the need for strict spatial alignment between augmented views, enabling spatial transformations, including flipping and scale changes, which are essential for NDC. This design also allows the model to learn cross-scale alignment, akin to how pathologists reason about the relationship between individual nuclei and their tissue context (Figure 2). To further enhance contextual learning, MUSE progressively expands the pretraining FoV up to 65.5K $\mu$m$^2$ (512×512 pixels). During fine-tuning, we propose a simple yet effective large-FoV semi-supervised strategy: labeled patches are expanded to 262.1K $\mu$m$^2$ (1024×1024 pixels), where supervised learning is applied to annotated regions, and pseudo-labeling is applied to surrounding unlabeled areas. Overall, the proposed MUSE framework effectively addresses key challenges in histopathological NDC. Extensive experiments on three widely used benchmarks demonstrate that MUSE-pretrained models not only significantly surpass existing supervised NDC methods but also outperform generic pathology foundation models even with smaller models and fewer samples. Our contributions are summarized as follows:
\begin{itemize}
\item We propose MUSE, a novel SSL approach specifically tailored for nucleus detection and classification. By introducing the novel Nucleus-based Local Self-distillation (NuLo) mechanism, MUSE enables flexible multi-scale local self-distillation. Combined with large-FoV pretraining, it allows the model to capture discriminative nucleus features.
\item Built around MUSE, we develop a complete framework that includes an encoder-decoder architecture, a pretraining strategy, and a downstream fine-tuning pipeline, allowing effective utilization of unlabeled data and LFoV across all training stages.
\item Extensive experiments demonstrate that our MUSE framework effectively addresses the key challenges of histopathological NDC. The resulting pretrained models not only surpass state-of-the-art supervised methods but also outperform generic pathology foundation models.
\end{itemize}

\section{Related Work}

\subsection{Nucleus Detection and Classification}
 
Current methods for NDC can be broadly categorized into map-based~\cite{hovernet,pointnu,smile,cellvit,cellvit++} and point-based methods~\cite{mcspatnet,ryu2023ocelot,dpap2pnet}.
Despite extensive research on model designs, these methods still follow the supervised learning paradigm, which relies on large-scale, fully annotated nucleus-level datasets.
On the other hand, semi-supervised learning has been explored to leverage unlabeled patches for better performance~\cite{bai2021novel,su2023dual}.
However, these methods typically rely on strong assumptions such as the cluster assumption~\cite{yang2022survey}, which limit the usage of large-scale heterogeneous pathology patches.
In this paper, we propose a novel SSL framework to leverage abundant unlabeled patches for better performance on NDC.

\subsection{Pathology Pretraining}

Self-supervised learning is an efficient strategy for learning generalizable representations from large-scale unlabeled data~\cite{misra2020self,chen2021empirical,ibot,mae,dino,assran2023self,dinov2}.
Recently, pathological pretraining methods have been extensively explored~\cite{wang2022transformer,azizi2023robust,ctranspath,uni,gigapath,chief}.
After being pretrained on large-scale unlabeled pathology patches, these methods substantially advance WSI-level tasks~\cite{chief}.
In this work, we experimentally shows that these methods still exhibit limited performance for nucleus representation.
To address this issue, we propose NuLo to achieve better local self-distillation based on matched nuclei.

\begin{figure}[t]
    \centering
    \includegraphics[width=0.99\linewidth]{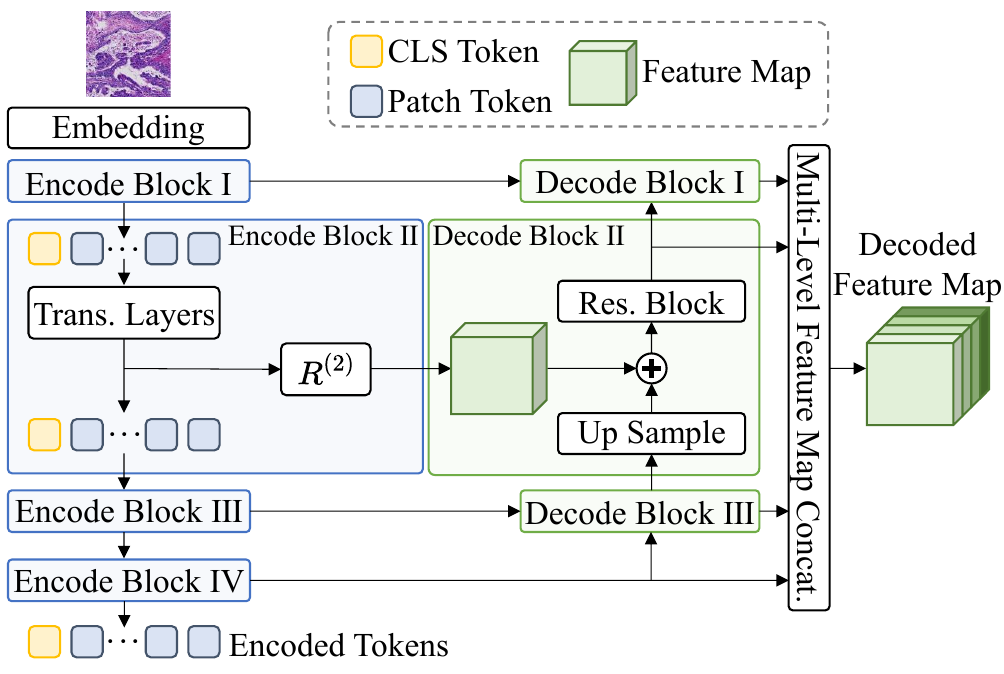}
    \caption{
    Illustration of the architecture.
    Given a patch, this architecture produces both image-level representation (CLS token) and high-resolution dense representations for multi-scale self-distillation. $R^{(2)}$ denotes the reassembly layer of the second encoder block.
    }
    \label{fig:Arch.}
\end{figure}

\begin{figure*}[t]
    \centering
    \includegraphics[width=0.99\textwidth]{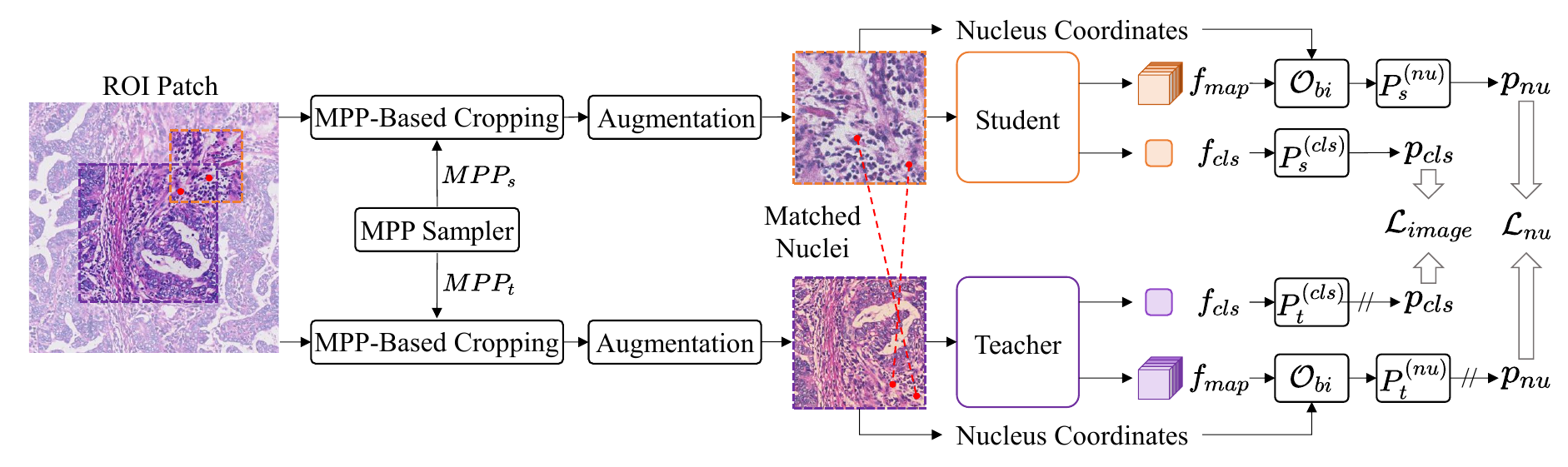}
    \caption{
    Illustration of MUSE.
    MPP-Based Cropping is first employed to generate paired views based on the ROI patch and random MPP.
    After data augmentation, we extract image-level representations (CLS tokens, $f_{cls}$) and dense representations (feature maps, $f_{map}$) of the paired views with teacher and student networks.
    MUSE minimizes two losses: 1) image-level self-distillation between CLS tokens and 2) nucleus-level self-distillation between features of matched nuclei.
    Specifically, nucleus features are interpolated from the feature maps based on their coordinates.
    }
    \label{fig:main_fig}
\end{figure*}

\section{Preliminaries}\label{sec:preliminaries}

\subsection{Nucleus Detection and Classification}

Nucleus detection and classification is defined as follows:
\begin{equation}
    \hat{C} = M(I), \hat{C} = \{[\hat{t^{(i)}}, \hat{x^{(i)}}, \hat{y^{(i)}}]^T\}_{i=1}^{N_p}
    \label{eq:def_cell_detection_classification}
\end{equation}
where $I$ is the input pathology patch, $M$ denotes the model, $N_p$ is the total number of predicted nuclei, $\hat{C}$ is the predicted set, $\hat{t^{(i)}}$, $\hat{x^{(i)}}$, and $\hat{y^{(i)}}$ are the predicted type, the coordinate of the x-axis and the y-axis of the $i$-th predicted nucleus, respectively.
In this work, we further define a simplified task, termed nucleus classification, to focus on evaluating nucleus representation performance:
\begin{equation}
    \{\hat{t^{(i)}}\}_{i=1}^{N_g} = M(I; \{[x^{(i)}, y^{(i)}]^T\}_{i=1}^{N_g}),
\end{equation}
where $N_g$ is the total number of ground truth nuclei, $x^{(i)}$ and $y^{(i)}$ are the ground truth x-axis and the y-axis coordinates of the $i$-th nucleus, respectively.

\subsection{Self-Distillation with No Labels}\label{sec:dino_update}

Self-Distillation with No Labels (DINO)~\cite{dino} employs a self-supervised teacher-student architecture to learn visual representations without labels. 
The student network $M_s$ and teacher network $M_t$ share identical architectures but maintain separate parameters.
Given two augmented views, denoted as $I^{(1)}$ and $I^{(2)}$, of an input image, the output of $M_s$ is required to match the output of $M_t$:
\begin{equation}
    \mathcal{L}_{image} = -P_t^{(cls)}(M_t(I^{(1)}))\log P_s^{(cls)}(M_s(I^{(2)})),
    \label{eq:dino_basic}
\end{equation}
where $P_t^{(cls)}$ and $P_s^{(cls)}$ denote the operator to compute output probability of $M_t$ and $M_s$, respectively, $\mathcal{L}_{image}$ is the cross-entropy loss at the image level.
In practice, the multi-crop strategy is applied to generate a set of views $V=\{I_{g}^{(1)}, I_{g}^{(2)}, I_{l}^{(i)}\}_{i=1}^{N_l}$ from an image to build a group of paired samples and encourage local-to-global learning, where $N_l$ is the number of local views, $I_g$ denote global view with large resolution, and $I_l$ denote local view with smaller resolution.
$M_s$ are updated with stochastic gradient descent to minimize $\mathcal{L}_{image}$.
$M_t$ are initialized as the same of $M_s$ and updated with an Exponential Moving Average (EMA) of $M_s$.
In this work, we mainly follow the framework of DINO and further introduce nucleus-level self-distillation to improve the representation of nuclei.

\section{Method}

As shown in Figure \ref{fig:Arch.}, we introduce a lightweight and flexible encoder-decoder backbone to extract features.
The framework of MUSE is illustrated in Figure \ref{fig:main_fig}.
First, MPP-based cropping is employed to obtain multi-scale paired views.
Second, multi-level representations of views are extracted with teacher and student networks.
Third, these representations are further utilized for image-level and nucleus-level self-distillation.
For downstream NDC tasks, the pretrained model is applied to the specific dataset with a large-FoV semi-supervised fine-tuning pipeline.

\subsection{Architecture}

The encoder-decoder framework has been extensively validated for NDC~\cite{hovernet, dpap2pnet}.
In this work, an encoder-decoder backbone based on Vision Transformer (ViT)~\cite{vit} is constructed.

The encoder is composed of a ViT and reassembly layers.
Specifically, we first extract multi-level encoded features $F_e = \{f_e^{(i)}\}_{i}^{N_e}$ from $N_e$ layers, where $f_e^{(i)}$ is the encoded feature from the $i$-th layer.
Subsequently, the reassembly layers $R(\cdot)$ are used to convert $F_e$ into the 2D feature maps required by the decoder: $F_e^\prime = \{R^{(i)}(f^{(i)}_e)\}_{i=1}^{N_e}$.
For ViT,$f_e^{(i)} \in \mathbb{R}^{(HW/p^2+1) \times c^{(i)}}$, where $p$ denotes the patch size, $c$ denotes the dimension of each token, $H$ denotes the height of the input image, and $W$ denotes the width of the input image.
The reassembly layer $R^{(i)}(\cdot)$ discards the CLS token, reassembles the sequence of tokens into a 2D feature map $\in \mathbb{R}^{c^{(i)} \times (H/p) \times (W/p)}$, adjusts the feature map to the target channel dimension via a 1×1 convolution, and finally resamples it to the target spatial size.

The decoder comprises a series of residual-block-based modules that progressively fuse feature maps from the encoder. 
Let $f_d^{(i)}$ denote the $i$-th decoded feature map, and let the full set of decoder outputs be $F_d=\{f_d^{(i)}\}_{i=1}^{N_e}$. Decoded feature maps are further mapped to the target channel dimension and spatial size, and then concatenated to form a unified feature map $f_{map}=Cat(F_d)$, which serves as the dense representation of the input image.
In addition, the CLS token output of the encoder, denoted as $f_{cls}$, is used as the image-level representation.

Following common practice in the encoder-decoder framework, $N_e$ is set to 4. 
$F_e^\prime$ is constructed from equally spaced transformer layers~\cite{dpt}.
This design serves as the default backbone configuration unless otherwise specified.

\subsection{MUSE}

\noindent \textbf{ROI Patches.}
WSIs typically exhibit gigapixel-scale dimensions, presenting challenges for constructing effective image pairs with partial overlap. To address this and generate suitable training samples, we introduce a sequential dual-cropping strategy: 1) Region of Interest (ROI) cropping and 2) multi-scale view cropping.
An initial crop operation is applied to the source WSI to isolate a relevant ROI.
A subsequent crop operation is performed on the extracted ROI to yield training samples.
In this work, a dataset comprising 483,627 ROI patches based on the Cancer Genome Atlas Program (TCGA)~\cite{liu2018integrated} is constructed, termed $\text{TCGA}_{\text{nu}}$. Nuclei coordinates are auto-detected. Please refer to the extended version in arXiv for details.

\noindent \textbf{MPP-Based Cropping.}
The physical scale of pathology images is critical for effective self-distillation in MUSE. 
To generate views of specified physical resolution, we propose a novel cropping method based on Microns-Per-Pixel (MPP).
The cropping operator $\mathcal{O}_{crop}$ is formally defined as:
\begin{equation}
    \{I_o, C_{o}\} = \mathcal{O}_{crop}(I_{e}, C_{e};MPP_e, MPP_o, R_o),
    \label{eq:build_sample_simplified}
\end{equation}
where $I_e$ is the input image, $C_e$ is the input nucleus coordinates, $I_o$ is the output image, $C_o$ is the output nucleus coordinates, $MPP_e$ is the input MPP, $MPP_o$ is the output MPP, and $R_o$ is the the pixels resolution of $I_o$.
$MPP_o$ is random generated by MPP Sampler.
$\mathcal{O}_{crop}$ crops a region of size $MPP_o / MPP_e\times R_o$ from $I_e$, then resizes it to $R_o$.

After cropping, $C_o$ is aligned to the local coordinate system of $I_o$, which makes it difficult to match cells in multiple output images from the same $I_s$.
We further introduce a coordinate-agnostic indexing scheme to uniquely identify each nucleus in $I_e$ for cross-view nucleus matching:
\begin{equation}
    \{I_o, C_{o}, K_o\} = \mathcal{O}_{crop}(I_{e}, C_{e}, \mathcal{O}_{in}(C_e));\cdots),
    \label{eq:build_sample}
\end{equation}
where $\mathcal{O}_{in}$ is the nucleus index construction operator, $K_o$ is the nucleus index list of $C_o$, and $\cdots$ denotes $[MPP_e, MPP_o, R_o]^T$.
For any two samples $\{I_o^{(1)}, C_{o}^{(1)}, K_o^{(1)}\}$ and $\{I_o^{(2)}, C_{o}^{(2)}, K_o^{(2)}\}$ derived from $I_e$, the nucleus-level match is efficiently established via the intersection of $K_o^{(1)}$ and $K_o^{(2)}$.

\begin{figure}[t]
    \centering
    \includegraphics[width=0.99\linewidth]{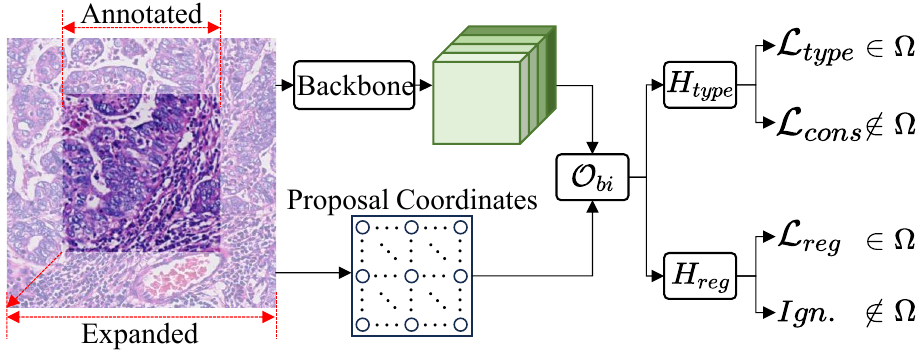}
    \caption{
    Illustration of the fine-tuning.
    Annotated samples are first extended to new samples with a larger field of view.
    The pretrained backbone is then used to extract feature maps, which are further utilized to obtain features at proposal coordinates.
    $Ign.$ denotes that the coordinate regression is not applied to unlabeled regions.
    }
    \label{fig:finetuning.}
\end{figure}

\noindent \textbf{Nucleus-Based Local Self-Distillation (NuLo).}
We introduce self-distillation at the nucleus level to encourage the model to distinguish nuclei in pathological images and to learn stable representations of nuclei across scales.
For any sample $\{I_o, C_o, K_o\}$, $f_{map}$ is extracted with the encoder-decoder architecture.
Each nucleus feature is then obtained via bilinear interpolation from $f_{map}$ based on coordinates:
$f_c = \mathcal{O}_{bi}(f_{uni}; [x_c, y_c]^T)$,
where $f_c$ denotes the feature vector of the nucleus, $\mathcal{O}_{bi}$ is the bilinear interpolation operator, and $[x_c, y_c]^T$ is the nucleus coordinates.
The set of nucleus features is denoted as $F_c = \{f_c\}^{C_o}$.
Given two views $\{I_o^{(1)}, C_o^{(1)}, K_o^{(1)}\}$ and $\{I_o^{(2)}, C_o^{(2)}, K_o^{(2)}\}$ derived from the same ROI patch, the corresponding nucleus features $F_c^{(1)}$ and $F_c^{(2)}$ are extracted through teacher and student networks, respectively.
Nucleus-level self-distillation loss ($\mathcal{L}_{nu}$) of each paired views is further defined as follows:
\begin{equation}
\begin{aligned}
    &\mathcal{L}_{nu} = \sum_{K_{cap}} -P_t^{(nu)}(F_c^{(1)}[K_{cap}]) \log P_s^{(nu)}(F_c^{(2)}[K_{cap}]), \\
    &s.t. \ K_{cap} = K_o^{(1)} \cap K_o^{(2)},
    \label{eq:cell-level_self-distillation}
\end{aligned}
\end{equation}
where $P_t^{(nu)}$ and $P_s^{(nu)}$ denote the operator to compute output probability of $F_c^{(1)}$ and $F_c^{(2)}$, $[\cdot]$ is the query operator based on index, $K_{cap}$ is the intersection set of $K_o^{(1)}$ and $K_o^{(2)}$.
Equation (\ref{eq:cell-level_self-distillation}) enables the student network to distinguish nuclei with morphological differences and learn cross-scale consistent representations of nuclei.

\noindent \textbf{Optimization.}
MUSE adopts the teacher-student network update method of DINO as described in Preliminaries.
Student network is optimized with both image-level ($\mathcal{L}_{image}$) and nucleus-level ($\mathcal{L}_{nu}$) self-distillation losses.
$\mathcal{L}_{image}$ is applied to the CLS token outputs of encoders to preserve global image representation learning.
Meanwhile, $\mathcal{L}_{nu}$ operates on the decoder outputs to enhance representation of nuclei.
The composite loss $\mathcal{L}_{MUSE}$ is further defined as:
\begin{equation}
    \mathcal{L}_{MUSE} = \lambda_{image} \mathcal{L}_{image} + \lambda_{nu} \mathcal{L}_{nu},
\end{equation}
where $\lambda_{image}$ and $\lambda_{nu}$ are the loss weights of $\mathcal{L}_{image}$ and $\mathcal{L}_{nu}$, respectively.
$\lambda_{image}$ and $\lambda_{nu}$ are set to 1 in experiments, unless otherwise mentioned.

\begin{table*}[h!]
  \centering
  \small
  \renewcommand\tabcolsep{3pt}
  \renewcommand{\arraystretch}{1.05}
  \begin{tabular}{ccccc|ccccccc}
    \toprule
    \multicolumn{1}{c}{\multirow{2}{*}{Method}} & \multicolumn{1}{c}{\multirow{2}{*}{Arch.}} & \multicolumn{1}{c}{\multirow{2}{*}{Params.}} & \multicolumn{1}{c}{\multirow{2}{*}{Dataset}} & \multicolumn{1}{c|}{\multirow{2}{*}{$N_s$}} & \multicolumn{3}{c}{Evaluation on 20x} & \multicolumn{3}{c}{Evaluation on 40x} & \multicolumn{1}{c}{\multirow{2}{*}{Overall}}\\
    & & & & & BRCA. & OCEL. & PUMA & BRCA. & OCEL. & PUMA\\
    \midrule
    \rowcolor{gray!12}
    \multicolumn{12}{c}{
        \textit{
        \textbf{Pretrained on Large-Scale General Datasets}
        }
    }\\ 

    DINO~\cite{dino} & ResNet-50 & 23M & IN-1k & 1M & 73.94 & 75.47 & 72.76 & 72.22 & 75.75 & 76.84 & 74.50 \\
    DINO~\cite{dino} & ViT-S/16 & 21M & IN-1k & 1M  & 77.85 & 81.76 & 78.36 & 82.33 & 80.37 & 78.80 & 79.91 \\
    DINO~\cite{dino} & ViT-B/16 & 85M & IN-1k & 1M  & 77.20 & 80.61 & 78.76 & 80.12 & 81.37 & 80.36 & 79.74 \\
    MAE~\cite{mae} & ViT-B/16 & 85M & IN-1k & 1M  & 76.78  & 78.46  & 77.83  & 77.56  & 81.38  & 77.27  & 78.21 \\
    iBOT~\cite{ibot} & ViT-S/16 & 21M & IN-22k & 14M  & 80.33  & 80.40  & 77.82  & 81.34  & 81.45  & 80.73  & 80.34 \\
    iBOT~\cite{ibot} & ViT-B/16 & 85M & IN-22k & 14M  & 79.57  & 82.48  & 77.93  & 82.48  & 82.17  & 79.07  & 80.62 \\
    DINOV2~\cite{dinov2} & ViT-S/14 & 21M & LVD-142M & 142M  & 82.39  & 80.13  & 79.66  & 81.72  & 79.13  & 81.55  & 80.76 \\
    DINOV2~\cite{dinov2} & ViT-B/14 & 86M & LVD-142M & 142M  & 84.39  & 81.66  & 80.66  & 85.02  & 79.37  & 82.93  & 82.34 \\

    \midrule
    \rowcolor{gray!12}
    \multicolumn{12}{c}{
        \textit{
        \textbf{Pretrained on Pathology Patches}
        }
    }\\ 

    MoCoV2~\cite{kang2023benchmarking} & ResNet-50 & 24M & TCGA & 19M & 80.71 & 82.17 & 79.60 & 79.27 & 82.20 & 79.90 & 80.64 \\
    DINO~\cite{kang2023benchmarking} & ViT-S/16 & 22M & TCGA & 19M & 83.63 & 85.30 & 81.92 & 84.30 & 84.44 & 81.13 & 83.45\\
    DINOV2~\cite{dinov2} & ViT-S/16 & 21M & $\text{TCGA}_{\text{nu}}$ & 484K & 81.71 & 83.65 & 79.08 & 83.34 & 83.04 & 79.13 & 81.66 \\
    DINOV2~\cite{dinov2} & ViT-B/16 & 86M & $\text{TCGA}_{\text{nu}}$ & 484K & 83.60 & 82.87 & 79.00 & 84.25 & 82.71 & 79.10 & 81.92\\
    CHIEF~\cite{chief} & Swin-T & 28M & $\text{custom}^*$ & 15M   & 79.52 & 82.77 & 77.69 & 79.77 & 81.42 & 76.11 & 79.55 \\
    CTransPath~\cite{ctranspath} & Swin-T & 28M & $\text{custom}^*$ & 15M & 80.12 & 82.92 & 77.83 & 79.87 & 81.07 & 78.15 & 79.99 \\
    CONCH~\cite{conch} & ViT-B/16 & 86M & $\text{custom}^*$ & $\text{1M}_{\text{(VL)}}$  & 86.09 & 87.12 & 82.52 & 87.93 & 86.32 & 84.80 & 85.80 \\
    UNI~\cite{uni} & ViT-L/16 & 300M & Mass-100k & 100M  & 87.12 & 87.72 & 83.04 & 88.95 & 87.26 & \underline{84.84} & 86.49 \\
    Prov-GigaPath~\cite{gigapath} & ViT-G/14 & 1.1B & $\text{custom}^*$ & 1.3B  & 86.99 & \textbf{88.13} & 83.46 & 88.00 & 86.03 & 83.04 & 85.94 \\

    \midrule

    \multicolumn{1}{c}{\multirow{3}{*}{\textbf{MUSE (ours)}}}
    & ResNet-50 & 86M & $\text{TCGA}_{\text{nu}}$ & 484K & 86.29 & 86.30 & 81.69 & 88.26 & 84.85 & 80.42 & 84.64 \\
    & ViT-S/16 & 31M & $\text{TCGA}_{\text{nu}}$ & 484K  & 86.40 & 86.21 & 83.09 & 88.56 & 86.40 & 80.79 & 85.24 \\
    & ViT-B/16 & 123M & $\text{TCGA}_{\text{nu}}$ & 484K  & 88.43 & 86.03 & 84.18 & 89.60 & 86.87 & 82.46 & 86.26 \\

    \midrule

    \multicolumn{1}{c}{\multirow{3}{*}{\textbf{LFoV-MUSE (ours)}}}
    & ResNet-50 & 86M & $\text{TCGA}_{\text{nu}}$ & 484K  & \underline{88.70} & \underline{87.87} & \underline{84.49} & \underline{89.74} & 85.17 & 82.62 & \underline{86.43} \\
    & ViT-S/16 & 31M & $\text{TCGA}_{\text{nu}}$ & 484K & 86.29 & 87.54 & 83.81 & 86.59 & \textbf{88.01} & 84.56 & 86.13 \\
    & ViT-B/16 & 123M & $\text{TCGA}_{\text{nu}}$ & 484K  & \textbf{89.29} & 87.05 & \textbf{84.84} & \textbf{90.26} & \underline{87.87} & \textbf{85.74} & \textbf{87.51} \\
    
    \bottomrule
  \end{tabular}
  \caption{Comparison of nucleus classification in ACC \% ($\uparrow$) for fine-tuning (FT). MUSE outperforms other pretraining methods in overall evaluations. The best results are highlighted in bold, and the second-best results are in underlined. $N_s$ is the number of samples in each dataset. $\text{custom}^*$ denotes mixed dataset. BRCA. and OCEL. denote BRCAM2C and OCELOT, respectively.}
  \label{tab:ssl_compare_ft}
\end{table*}

\noindent \textbf{LFoV Pretraining.}
Following the common practice of DINO, MUSE sets the pixel resolutions of global views and local views to 224 and 96, respectively. 
In addition, we further extend the pixel resolutions of global and local views to 512 and 208 to explore the impact of incorporating more tissue context on nucleus representation.

\subsection{Downstream Fine-Tuning}

To transfer the MUSE-pretrained model to NDC tasks, we propose a novel fine-tuning pipeline (Figure \ref{fig:finetuning.}).
We first naturally expand the small FOV samples to include more tissue context, and then employ the point-based method to regress nucleus coordinates and predict nucleus types based on independent heads $H_{reg}$ and $H_{type}$, respectively.

\noindent \textbf{LFoV Samples.}
While the framework can be generalized to arbitrary region shapes, we primarily discuss the square-shaped annotated regions, which align with the predominant structure of most existing datasets~\cite{puma, ryu2023ocelot, mcspatnet}.
Training samples, comprising cropped WSI regions with corresponding nucleus annotations, are located by their top-left corner coordinates $[x_a, y_a]^T$ and side length of the annotation region $r_a$.
Furthermore, the LFoV samples are generated by extending these annotated regions to incorporate surrounding unlabeled tissue areas:
\begin{equation}
    \begin{aligned}
        &x_a^\prime = x_a - \mathcal{O}_{sample}(0, r_a^\prime - r_a), \\
        &y_a^\prime = y_a - \mathcal{O}_{sample}(0, r_a^\prime - r_a), \\
        &s.t. \  r^\prime_a \geq r_a,
    \end{aligned}
    \label{eq:build_lfov}
\end{equation}
where $\mathcal{O}_{sample}$ is a random sampling operator, and $[x_a^\prime, y_a^\prime]^T$ and $r_a^\prime$ are the top-left corner coordinates and side length of the LFoV sample, respectively.
In Equation (\ref{eq:build_lfov}), region offsets are randomly sampled from the range $[0, r_a^\prime - r_a]$.
These samples permit the annotated region to appear at any position within the sample and provide expanded contextual tissue information beyond the annotation region.

\noindent \textbf{Semi-Supervised Fine-Tuning.}
For LFoV samples containing both annotated and unannotated regions, we naturally introduce a semi-supervised fine-tuning: 1) the coordinate regression loss $\mathcal{L}_{reg}$ and classification loss $\mathcal{L}_{type}$ of the annotated region, and 2) the consistency prediction regularization term $\mathcal{L}_{cons}$ of the unannotated region:
\begin{equation}
    \mathcal{L}_{ft} = \lambda_{reg}\mathcal{L}_{reg} + \lambda_{type}\mathcal{L}_{type} + \lambda_{cons}\mathcal{L}_{cons},
\end{equation}
where $\lambda_{reg}$, $\lambda_{cls}$, and $\lambda_{cons}$ are loss weights, and $\mathcal{L}_{ft}$ is the total loss for fine-tuning.
Specifically, we perform nucleus detection and classification on the entire LFoV sample, then split all predictions into two groups based on the annotated region $\Omega$.
Following common practice~\cite{dpap2pnet}, Mean Squared Error (MSE) and cross-entropy losses are employed for regression and classification, respectively.
Both $\mathcal{L}_{reg}$ and $\mathcal{L}_{type}$ are computed by ground truth annotations and corresponding predictions strictly within $\Omega$.
For unlabeled area, we first generate pseudo-labels with predicted probabilities, and then filter proposal points with prediction confidence above a specified threshold for $\mathcal{L}_{cons}$.

\section{Experiments}

\subsection{Experiment Settings}

\noindent \textbf{Dataset \& Metrics.}
BRCAM2C~\cite{mcspatnet}, OCELOT~\cite{ryu2023ocelot}, and PUMA~\cite{puma} are used to evaluate the performance of models on various tissues.
Following common practice~\cite{dino, dinov2} for evaluating pretrained models, we report Accuracy (ACC) of K-Nearest Neighbors (KNN), linear probing (LIN), and end-to-end fine-tuning (FT) for the nucleus classification task.
For NDC, we follow other SOTA methods~\cite {dpap2pnet} to evaluate models with F1 score.

\noindent \textbf{Baselines.}
We compare MUSE against strong SSL baselines pre-trained on large-scale general datasets, including iBOT~\cite{ibot},  MAE~\cite{mae}, DINO~\cite{dinov2}, and DINOv2~\cite{dinov2}.
Furthermore, we compare MUSE with the SOTA pathology foundation models, including PathBench~\cite{kang2023benchmarking}, CHIEF~\cite{chief}, CTransPath~\cite{ctranspath}, CONCH~\cite{conch}, Prov-GigaPath~\cite{gigapath}, and UNI~\cite{uni}.
Besides, we also compare fine-tuned MUSE-pretrained models with SOTA NDC methods, including MCSpatNet~\cite{mcspatnet}, PointNu-Net~\cite{pointnu}, CellViT~\cite{cellvit}, SMILE~\cite{smile}, SENC~\cite{senc}, CGT~\cite{cgt}, and DPA-P2PNet~\cite{dpap2pnet}.

\begin{table*}[!ht]
  \centering
  \small
  \renewcommand\tabcolsep{2pt}
  \renewcommand{\arraystretch}{1.05}
  \begin{tabular}{ccc|cccccccccccccc}
    \toprule
    \multicolumn{1}{c}{\multirow{2}{*}{Method}} & \multicolumn{1}{c}{\multirow{2}{*}{Arch.}} & \multicolumn{1}{c|}{\multirow{2}{*}{Dataset}} & \multicolumn{2}{c}{BRCA. (20x)} & \multicolumn{2}{c}{OCEL. (20x)} & \multicolumn{2}{c}{PUMA (20x)} & \multicolumn{2}{c}{BRCA. (40x)} & \multicolumn{2}{c}{OCEL. (40x)} & \multicolumn{2}{c}{PUMA (40x)} & \multicolumn{2}{c}{Overall}\\
    & & & KNN & LIN & KNN & LIN  & KNN & LIN  & KNN & LIN  & KNN & LIN  & KNN & LIN  & KNN & LIN \\
    \midrule
    \rowcolor{gray!17}
    \multicolumn{17}{c}{
        \textit{
        \textbf{Pretrained on Large-Scale General Datasets}
        }
    }\\ 

    DINO & ResNet-50 & IN-1k & 76.29 & 75.91 & 73.00 & 78.39 & 74.04 & 75.14 & 75.37 & 72.06 & 70.72 & 73.23 & 74.06 & 76.71 & 73.91 & 75.24 \\
    DINO & ViT-S/16 & IN-1k & 77.25 & 69.86 & 71.94 & 74.52 & 70.22 & 73.42 & 77.78 & 73.89 & 70.72 & 73.88 & 71.29 & 75.19 & 73.20 & 73.46\\
    DINO & ViT-B/16 & IN-1k & 78.03 & 77.09 & 73.11 & 76.83 & 71.40 & 76.80 & 76.28 & 74.59 & 72.30 & 76.92 & 72.86 & 78.43 & 74.00 & 76.78\\
    MAE & ViT-B/16 & IN-1k & 65.45 & 73.54 & 67.04 & 77.34 & 61.79 & 76.16 & 67.45 & 71.92 & 66.18 & 76.17 & 64.14 & 76.70 & 65.34 & 75.31\\
    iBOT & ViT-S/16 & IN-22k & 77.67 & 78.61 & 74.70 & 78.01 & 71.32 & 76.59 & 78.00 & 78.44 & 72.42 & 76.58 & 71.18 & 76.57 & 74.21 & 77.47\\
    iBOT & ViT-B/16 & IN-22k & 79.06 & 76.20 & 76.66 & 77.54 & 71.09 & 76.35 & 80.63 & 79.97 & 74.86 & 78.85 & 73.34 & 78.21 & 75.94 & 77.85\\
    DINOV2 & ViT-S/14 & LVD-142M & 80.27 & 78.49 & 76.77 & 76.79 & 69.55 & 75.93 & 78.80 & 74.95 & 77.35 & 78.37 & 73.26 & 78.06 & 76.00 & 77.10\\
    DINOV2 & ViT-B/14 & LVD-142M & 79.11 & 78.96 & 75.25 & 78.22 & 68.81 & 76.09 & 80.12 & 74.68 & 76.06 & 79.28 & 73.50 & 79.45 & 75.48 & 77.78\\

    \midrule
    \rowcolor{gray!17}
    \multicolumn{17}{c}{
        \textit{
        \textbf{Pretrained on Pathology Patches}
        }
    }\\ 

    MoCoV2 & ResNet-50 & TCGA & 79.37 & 81.90 & 76.83 & 79.31 & 74.52 & 78.92 & 78.23 & 80.65 & 78.03 & 81.46 & 75.48 & 79.55 & 77.08 & 80.30 \\
    DINO & ViT-S/16 & TCGA & 84.31 & 80.85 & 82.06 & 83.61 & 76.69 & 79.89 & 80.55 & 82.15 & 78.42 & 82.58 & 76.20 & 80.61 & 79.70 & 81.61\\
    DINOV2 & ViT-S/16 & $\text{TCGA}_{\text{nu}}$ & 78.56 & 80.61 & 75.44 & 82.17 & 71.94 & 77.53 & 77.08 & 81.85 & 73.98 & 80.87 & 71.74 & 77.85 & 74.79 & 80.15\\
    DINOV2 & ViT-B/16 & $\text{TCGA}_{\text{nu}}$ & 77.90 & 82.21 & 75.22 & 83.25 & 72.06 & 78.26 & 77.73 & 82.35 & 74.13 & 81.98 & 71.17 & 78.88 & 74.70 & 81.16\\
    CHIEF & Swin-T & $\text{custom}^*$ & 78.39 & 78.89 & 80.21 & 81.88 & 74.17 & 76.90 & 75.71 & 78.05 & 73.76 & 78.13 & 73.73 & 75.49 & 75.99 & 78.22\\
    CTransPath & Swin-T & $\text{custom}^*$ & 80.39 & 78.80 & 80.07 & 81.47 & 74.89 & 76.56 & 77.73 & 77.43 & 73.77 & 77.52 & 73.62 & 76.13 & 76.75 & 77.98\\
    CONCH & ViT-B/16 & $\text{custom}^*$ & 86.68 & 85.13 & \textbf{88.08} & 86.41 & 81.95 & 82.02 & 86.71 & 86.20 & \textbf{86.00} & 83.80 & \underline{83.05} & \underline{83.66} & 85.41 & 84.54\\
    UNI & ViT-L/16 & Mass-100k & 87.65 & 86.99 & 87.35 & 86.17 & \textbf{82.21} & 82.37 & \underline{88.82} & 88.64 & \underline{85.90} & 85.72 & 81.95 & 83.49 & \underline{85.65} & 85.56\\
    Prov-GigaPath & ViT-G/14 & $\text{custom}^*$ & 86.44 & 85.99 & 86.99 & \textbf{87.50} & 80.24 & 81.79 & 86.49 & 87.66 & 83.59 & 83.80 & 79.90 & 81.83 & 83.94 & 84.76\\

    \midrule

    \multicolumn{1}{c}{\multirow{3}{*}{\textbf{MUSE}}}
    & ResNet-50 & $\text{TCGA}_{\text{nu}}$ & 88.37 & 88.14 & 85.51 & 85.57 & 81.21 & 81.53 & 85.78 & 87.39 & 83.49 & 83.65 & 78.60 & 80.64 & 83.82 & 84.49 \\
    & ViT-S/16 & $\text{TCGA}_{\text{nu}}$ & 86.88 & 87.79 & 86.13 & 85.42 & 80.00 & 81.34 & 87.67 & \underline{89.66} & 85.45 & 85.20 & 79.71 & 80.17 & 84.31 & 84.93\\
    & ViT-B/16 & $\text{TCGA}_{\text{nu}}$ & 87.56 & \underline{89.60} & 85.90 & 85.82 & 81.26 & 83.29 & 88.11 & 88.86 & 85.55 & 85.57 & 81.19 & 82.48 & 84.93 & 85.94\\

    \midrule

    \multicolumn{1}{c}{\multirow{3}{*}{\textbf{LFoV-MUSE}}}
    & ResNet-50 & $\text{TCGA}_{\text{nu}}$ & \textbf{89.53} & \textbf{90.18} & 86.21 & 86.19 & \textbf{82.21} & 83.85 & 87.44 & 88.86 & 85.18 & 85.78 & 79.88 & 82.76 & 85.07 & \underline{86.27} \\
    & ViT-S/16 & $\text{TCGA}_{\text{nu}}$ & 85.47 & 87.06 & 84.17 & \underline{87.21} & 79.15 & \underline{84.22} & 86.00 & 86.63 & 84.95 & \textbf{86.57} & 79.82 & 83.53 & 83.26 & 85.87\\
    & ViT-B/16 & $\text{TCGA}_{\text{nu}}$ & \underline{89.03} & 89.20 & \underline{87.38} & 86.10 & 81.11 & \textbf{84.36} & \textbf{88.93} & \textbf{90.18} & 85.52 & \underline{86.43} & \textbf{83.16} & \textbf{85.12} & \textbf{85.86} & \textbf{86.90}\\
    
    \bottomrule
  \end{tabular}
  \caption{Comparison of nucleus classification in ACC \% ($\uparrow$) for K-Nearest Neighbor (KNN) and Linear Probing (LIN). MUSE outperforms other pretraining methods in overall evaluations. The best results are highlighted in bold, and the second-best results are in underlined. $\text{custom}^*$ denotes mixed datasets. BRCA. and OCEL. denote BRCAM2C and OCELOT, respectively.}
  \label{tab:ssl_compare_knn_lin}
\end{table*}

\noindent \textbf{Implementation.} Please refer to the extended version in arXiv for detailed hyper-parameters.

\subsection{Main Results}

\noindent \textbf{Nucleus Classification.}
Table \ref{tab:ssl_compare_ft} and Table \ref{tab:ssl_compare_knn_lin} report the results of nucleus classification on multi-tissue and multi-magnification datasets.
Pathology pretraining significantly improves nucleus representation performance compared to models pretrained on general datasets.
Furthermore, MUSE achieves better data efficiency and overall nucleus classification performance compared to existing methods.
Specifically, ViT-B pre-trained with MUSE outperforms CONCH, which has the same encoder, in ACC \% by 1.4 in LIN and 0.5 in FT.
As CONCH uses over 1 million image-text pairs (versus about 0.5 million patches for MUSE), CONCH shows a marginal advantage over MUSE in KNN.

Furthermore, we conducted pretraining and inference of MUSE with a larger field of view (LFoV-MUSE).
Compared to MUSE, ViT-S and ViT-B pretrained with LFoV-MUSE exhibit improvements in FT ACC \% of 0.9 and 1.3, respectively.
ViT-B pretrained with LFoV-MUSE also outperforms all other SOTA foundation models in KNN, LIN, and FT evaluations.
Specifically, it outperforms CONCH by 0.5, 2.4, and 1.7 in KNN, LIN, and FT ACC \%, respectively.
Remarkably, it also exceeds the performance of much larger models: outperforming UNI (2.4× parameters) by 0.2, 1.3, and 1.0 in KNN, LIN, and FT ACC \%, respectively, and surpassing Prov-GigaPath (8.9× parameters) by 1.9, 2.1, and 1.6 in KNN, LIN, and FT ACC \%, respectively.

For a fairer comparison, we pretrained models with DINOV2 on $\text{TCGA}_{\text{nu}}$.
While ViT-S benefits from pre-training on $\text{TCGA}_{\text{nu}}$ compared to LVD-142M, ViT-B exhibits decreased performance, suggesting that the data size of $\text{TCGA}_{\text{nu}}$ is not enough for large-parameter model pretraining with DINOV2.
In contrast, MUSE shows better data efficiency with the same dataset.

\begin{table*}
  \centering
  \small
 \renewcommand\tabcolsep{4pt}
 \renewcommand{\arraystretch}{1.05}
 \begin{tabular}{c|cccc|ccc|cccc}
    \toprule

    \multicolumn{1}{c|}{\multirow{2}{*}{Method}} & 
    \multicolumn{4}{c|}{BRCAM2C} & 
    \multicolumn{3}{c|}{OCELOT} & 
    \multicolumn{4}{c}{PUMA}\\
    & $F^{Lym.}$ & $F^{Tum.}$ & $F^{Oth.}$ & $F^{Avg.}$ & $F^{Tum.}$ & $F^{Oth.}$ & $F^{Avg.}$ & $F^{Lym.}$ & $F^{Tum.}$ & $F^{Oth.}$ & $F^{Avg.}$ \\

    \midrule

    MCSpatNet~\cite{mcspatnet} & 63.15 & 78.56 & 54.66 & 65.46 & 68.60 & 59.99 & 64.29 & 78.25 & 82.05 & 51.54 & 70.61\\
    PointNu-Net~\cite{pointnu} & \underline{71.51} & 76.02 & 51.95 & 66.50 & 66.72 & 56.96 & 61.84 & 76.31 & 79.57 & 52.68 & 69.52 \\
    SMILE~\cite{smile} & \textbf{72.59} & \underline{79.61} & 51.06 & \underline{67.75} & 66.99 & 60.10 & 63.55 & \underline{80.35} & 77.54 & 52.52 & 70.14 \\

    SENC~\cite{senc} & 57.94 & 76.50 & 49.42 & 61.29 & 70.02 & \underline{62.08} & \underline{66.05} & 77.38 & 81.51 & 54.38 & 71.09\\
    CGT~\cite{cgt} & 56.42 & 75.98 & 50.44 & 60.95 & 68.77 & 61.30 & 65.03 & 76.55 & 79.66 & 54.20 & 70.14\\
    CellViT~\cite{cellvit} & 67.20 & 78.20 & 51.81 & 65.73 & 67.36 & 60.22 & 63.79 & 79.07 & 81.16 & \underline{57.96} & \underline{72.73} \\
    DPA-P2PNet~\cite{dpap2pnet} & 59.65 & 77.26 & \underline{55.26} & 64.06 & \underline{70.07} & 59.92 & 64.99 & 76.80 & \underline{81.87} & 54.04 & 70.90 \\

    \midrule

    \textbf{LFoV-MUSE [ViT-B/16] (ours)} & 70.27 & \textbf{83.48} & \textbf{61.36} & \textbf{71.70} & \textbf{76.37} & \textbf{70.20} & \textbf{73.29} & \textbf{80.75} & \textbf{84.53} & \textbf{63.62} & \textbf{76.30} \\
    
    \bottomrule
  \end{tabular}
  \caption{
  Comparison of nucleus detection and classification in F1-score ($\uparrow$).
  Pretrained ViT-B/16 with LFoV-MUSE significantly outperforms other methods. The best results are highlighted in bold, and the second-best results are in underlined.}
  \label{tab:finetune}
\end{table*}

MUSE can be adapted to models with various encoder architectures.
Table \ref{tab:ssl_compare_ft} and Table \ref{tab:ssl_compare_knn_lin} shows that MUSE also significantly outperforms other methods when using ResNet-50~\cite{he2016deep} as the encoder.
The experiments based on different encoders verify the flexibility of MUSE.

\noindent \textbf{Nucleus Detection and Classification.}
Table \ref{tab:finetune} reports the results of nucleus detection and classification.
After pretraining with MUSE, simple downstream fine-tuning yields substantially better nucleus classification performance compared to both map-based and point-based SOTA methods.
Specifically, ViT-B pretrained with LFoV-MUSE outperforms the best SOTA methods by an average F1-score margin of 3.95, 7.24, and 3.57 on the BRCAM2C, OCELOT, and PUMA datasets, respectively.
Importantly, unlike previous work that requires spatial nucleus density statistics~\cite{mcspatnet} or nucleus graph construction~\cite{cgt}, our method achieves superior results with a much simpler task-adaptive pipeline.

\subsection{Ablation Studies}

All experiments are conducted with ViT-B/16 and the small field-of-view, unless otherwise mentioned.

\noindent \textbf{Decoder Pretraining.}
For models without a decoder, NuLo is applied to the last layer output of the encoder.
As shown in Table \ref{tab:ablation_multi-level-feature}, although the baseline without a decoder still outperforms DINOV2 pretrained with $\text{TCGA}_{\text{Nu}}$, there is a noticeable drop in performance.
Specifically, the model employing both the decoder and multi-level hybrid nucleus representations outperforms the baseline by 3.01, 1.98, and 1.79 in KNN, LIN, and FT ACC \%, respectively. 
Furthermore, removing multi-level hybrid nucleus representations also leads to a decline in performance.
%





    

\begin{table}
  \centering
  \small
 \renewcommand\tabcolsep{5pt}
 \renewcommand{\arraystretch}{1.05}
 \begin{tabular}{cc|ccc}
    \toprule

    Decoder & Multi-Level Context & KNN & LIN & FT \\


    \midrule

    \ding{55} & \ding{55} & 81.92 & 83.96 & 84.47\\
    \ding{51} & \ding{55} & 83.99 & 85.01 & 86.19\\
    \ding{51} & \ding{51} & 84.93 & 85.94 & 86.26\\
    
    \bottomrule
  \end{tabular}
  \caption{
  Ablation study of encoder-decoder framework in ACC \% ($\uparrow$).
  Introducing a decoder and multi-level context to the pretraining effectively enhances model performance.
  \label{tab:ablation_multi-level-feature}
  }
\end{table}

\begin{table}
  \centering
  \small
 \renewcommand\tabcolsep{2pt}
 \renewcommand{\arraystretch}{1.05}
 \begin{tabular}{cc|cccccc}
    \toprule

    Global View & Local View & \multicolumn{3}{c}{20x Evaluation} & \multicolumn{3}{c}{40x Evaluation}\\
    
    [Min, Max] & [Min, Max] & KNN & LIN & FT & KNN & LIN & FT \\
    
    \midrule
    $\text{[20×, 20×]}$ & $\text{[20×, 20×]}$ & 82.11 & 83.77 & 84.74 & 77.55 & 81.01 & 82.79\\
    $\text{[40×, 40×]}$ & $\text{[40×, 40×]}$ & 79.11 & 81.08 & 82.89 & 84.21 & 85.65 & 85.47 \\
    $\text{[20×, 40×]}$ & $\text{[20×, 40×]}$ & 84.91 & 86.24 & 86.21 & 84.95 & 85.64 & 86.31 \\
    \bottomrule
  \end{tabular}
  \caption{
  Ablation study of multi-scale patching in ACC \% ($\uparrow$).
  Pretraining with Multi-scale patching enables the model to adapt to multiple magnifications.
  }
  \label{tab:ablation_multi_scale_patching}
\end{table}

\noindent \textbf{Multi-Scale Patching.}
MPP-based Cropping support precise multi-scale patching based on physical resolution.
Table \ref{tab:ablation_multi_scale_patching} reports the ablation study of multi-scale patching.
Pretraining at a fixed magnification leads to a significant performance drop at other magnifications.
In contrast, pretraining with multi-scale patching enables the model to adapt to different magnifications and improves performance across all magnifications.
These results show that multi-scale patching enables MUSE to learn more robust nucleus representations.

\noindent \textbf{Pretraining Losses.}
Table \ref{tab:ablation_pretrain_loss} reports the ablation study of pretraining losses.
Building up on the baseline with $\mathcal{L}_{image}$, further introducing $\mathcal{L}_{nucleus}$ leads to better performance of 3.56, 3.42, and 3.92 in KNN, LIN, and FT ACC \%, respectively.
These results verify the critical role of nucleus-level contrastive learning for better nucleus representation.

\noindent \textbf{Fine-Tuning.}
Table \ref{tab:ablation_fine-tuning} reports the ablation study of fine-tuning on OCELOT.
Notably, even without introducing LFoV or $\mathcal{L}_{type}$, pretrained ViT-B by LFoV-MUSE already outperforms other SOTA methods, highlighting the efficiency of MUSE.
Expanding samples to LFoV samples further enriches tissue-level context, resulting in a further average classification F1 increase of 1.63.
Besides, the addition of the consistency regularization term $\mathcal{L}_{cons}$ yields an overall improvement of 4.29 relative to the baseline.
These results shows the importance of tissue-level context, and verify that the consistency prediction constraint based on unlabeled regions further enhances the generalization of models.

\begin{table}
  \centering
  \small
  \renewcommand\tabcolsep{5pt}
  \renewcommand{\arraystretch}{1.05}
  \begin{tabular}{cc|ccc}
    \toprule

    $\mathcal{L}_{image}$ & $\mathcal{L}_{nuclei}$ & KNN & LIN & FT \\

    \midrule

    \ding{51} & \ding{55} & 81.37 & 82.52 & 82.34\\
    \ding{55} & \ding{51} & 84.14 & 84.31 & 84.19\\
    \ding{51} & \ding{51} & 84.93 & 85.94 & 86.26\\
    
    \bottomrule
  \end{tabular}
  \caption{
  Ablation study of the pretraining losses in ACC \% ($\uparrow$).
  $\mathcal{L}_{nuclei}$ highly improves the representation of nuclei.
  \label{tab:ablation_pretrain_loss}
  }
\end{table}

\begin{table}
  \centering
  \small
 \renewcommand\tabcolsep{5pt}
 \renewcommand{\arraystretch}{1.05}
 \begin{tabular}{cc|cccc}
    \toprule

    LFoV & $\mathcal{L}_{cons}$ & $F^{Tum.}$ & $F^{Oth.}$ & $F^{Avg.}$ \\

    \midrule

    \ding{55} & \ding{55} & 73.48 & 64.52 & 69.00 \\ 
    \ding{51} & \ding{55} & 74.60 & 66.66 & 70.63 \\ 
    \ding{51} & \ding{51} & 76.37 & 70.20 & 73.29 \\ 
    
    \bottomrule
  \end{tabular}
  \caption{
  Ablation study of LFoV and $\mathcal{L}_{cons}$ on OCELOT in F1-score ($\uparrow$). Both improve the performance.
  }
  \label{tab:ablation_fine-tuning}
\end{table}

\begin{table*}[t]
  \centering
 \renewcommand\tabcolsep{3pt}
 \renewcommand{\arraystretch}{1.2}
 \resizebox{\textwidth}{!}{
 \begin{tabular}{cc|cc|cc|ccccccccccccccccccccc}
    \toprule

    \multicolumn{1}{c}{\multirow{2}{*}{Decoder}} & \multicolumn{1}{c|}{\multirow{2}{*}{Multi-Level Context}} & \multicolumn{1}{c}{\multirow{2}{*}{Global View}} & \multicolumn{1}{c|}{\multirow{2}{*}{Local View}} & \multicolumn{1}{c}{\multirow{2}{*}{$\mathcal{L}_{image}$}} & \multicolumn{1}{c|}{\multirow{2}{*}{$\mathcal{L}_{nu}$}} & \multicolumn{3}{c}{BRCAM2C (20x)} & \multicolumn{3}{c}{OCELOT (20x)} & \multicolumn{3}{c}{PUMA (20x)} & \multicolumn{3}{c}{BRCAM2C (40x)} & \multicolumn{3}{c}{OCELOT (40x)} & \multicolumn{3}{c}{PUMA (40x)} & \multicolumn{3}{c}{Overall}\\
    &  & & & & & KNN & LIN & FT & KNN & LIN & FT & KNN & LIN & FT & KNN & LIN & FT & KNN & LIN & FT & KNN & LIN & FT & KNN & LIN & FT\\

    \midrule
    \rowcolor{gray!27}
    \multicolumn{27}{c}{
        \textit{
        \textbf{Decoder Pretraining}
        }
    }\\ 

    \ding{55} & \ding{55} & $\text{[20×, 40×]}$ & $\text{[20×, 40×]}$ & \ding{51} & \ding{51} & 84.65 & 86.92 & 86.78 & 84.12 & 84.57 & 85.01 & 76.60 & 78.72 & 79.77 & 86.55 & 87.74 & 88.23 & 82.54 & 83.99 & 85.96 & 77.08 & 81.84 & 81.08 & 81.92 & 83.96 & 84.47\\
    \ding{51} & \ding{55} & $\text{[20×, 40×]}$ & $\text{[20×, 40×]}$ & \ding{51} & \ding{51} & 87.89 & 88.68 & 88.34 & 86.00 & 86.60 & 87.29 & 79.65 & 80.61 & 83.50 & 88.36 & 88.98 & 90.46 & 85.10 & 85.93 & 85.68 & 76.93 & 79.26 & 81.89 & 83.99 & 85.01 & 86.19\\

    \midrule
    \rowcolor{gray!27}
    \multicolumn{27}{c}{
        \textit{
        \textbf{Multi-Scale Patching}
        }
    }\\ 

    \ding{51} & \ding{51} & $\text{[20×, 20×]}$ & $\text{[20×, 20×]}$ & \ding{51} & \ding{51} & 83.75 & 86.60 & 87.06 & 85.70 & 84.52 & 86.11 & 76.88 & 80.19 & 81.04 & 79.86 & 82.21 & 84.21 & 78.24 & 81.18 & 84.16 & 74.54 & 79.63 & 79.99 & 79.83 & 82.39 & 83.76\\
    \ding{51} & \ding{51} & $\text{[40×, 40×]}$ & $\text{[40×, 40×]}$ & \ding{51} & \ding{51} & 80.82 & 81.27 & 82.76 & 81.11 & 81.55 & 84.47 & 75.38 & 80.41 & 81.45 & 87.58 & 88.02 & 88.98 & 83.71 & 85.53 & 86.36 & 81.34 & 83.40 & 81.08 & 81.66 & 83.36 & 84.18\\

    \midrule
    \rowcolor{gray!27}
    \multicolumn{27}{c}{
        \textit{
        \textbf{Pretraining Losses}
        }
    }\\

    \ding{51} & \ding{51} & $\text{[20×, 40×]}$ & $\text{[20×, 40×]}$ & \ding{51} & \ding{55} & 85.08 & 85.03 & 84.38 & 83.52 & 83.77 & 84.92 & 76.00 & 80.69 & 79.03 & 85.71 & 81.28 & 81.85 & 81.41 & 83.03 & 84.45 & 76.48 & 81.35 & 79.43 & 81.37 & 82.52 & 82.34\\
    \ding{51} & \ding{51} & $\text{[20×, 40×]}$ & $\text{[20×, 40×]}$ & \ding{55} & \ding{51} & 86.64 & 86.18 & 85.89 & 86.52 & 84.87 & 85.46 & 79.93 & 82.39 & 81.26 & 87.82 & 88.01 & 87.36 & 84.33 & 84.39 & 85.58 & 79.59 & 80.04 & 79.57 & 84.14 & 84.31 & 84.19\\

    \midrule
    \rowcolor{gray!27}
    \multicolumn{27}{c}{
        \textit{
        \textbf{Multi-Scale Dense Self-Distillation}
        }
    }\\

    \ding{51} & \ding{51} & $\text{[20×, 40×]}$ & $\text{[20×, 40×]}$ & \ding{51} & \ding{51} & 87.56 & 89.60 & 88.43 & 85.90 & 85.82 & 86.03 & 81.26 & 83.29 & 84.18 & 88.11 & 88.86 & 89.60 & 85.55 & 85.57 & 86.87 & 81.19 & 82.48 & 82.46 & 84.93 & 85.94 & 86.26\\
    
    \bottomrule
  \end{tabular}
  }
  \caption{\textbf{Detailed ablation studies of MUSE in ACC \% ($\uparrow$).} Introducing decoder pertaining, multi-scale patching, and Nucleus-Based Local Self-Distillation effectively enhances model performance.}
  \label{tab:ablation_fully_pretrain}
\end{table*}

\section{Conclusion}

In this work, we propose MUSE, a novel SSL method specifically tailored for nucleus detection and classification.
Built around MUSE, we develop a complete framework comprising an encoder-decoder architecture, a pretraining strategy, and a downstream fine-tuning pipeline, enabling effective utilization of unlabeled data and LFoV across all training stages.
Extensive experiments demonstrate that MUSE-pretrained models not only significantly surpass existing supervised NDC methods but also outperform generic pathology foundation models even with smaller models and fewer samples.
This work highlights the critical role of task-specific pretraining in nucleus-level dense prediction tasks, provides an effective and scalable solution, and paves the way toward general-purpose NDC.

\newpage

\section{Acknowledgments}
This work was supported by DAMO Academy, Alibaba Group, and the National Key R\&D Program of China (No.2024YFF0728900).

\appendix

\section{Appendix A: Additional Results}

\subsection{Detailed Results of Ablation Studies}

The complete results of the pretraining ablation studies in the main text are presented in Table \ref{tab:ablation_fully_pretrain}. Built around MUSE, we propose several new modules, including decoder pretraining, multi-scale patching, and NuLo. 
All of these modules show improvements over the baseline in multiple scales and tissue types evaluations.

\subsection{Visualization}

\begin{figure}[t]
    \centering
    \includegraphics[width=0.99\linewidth]{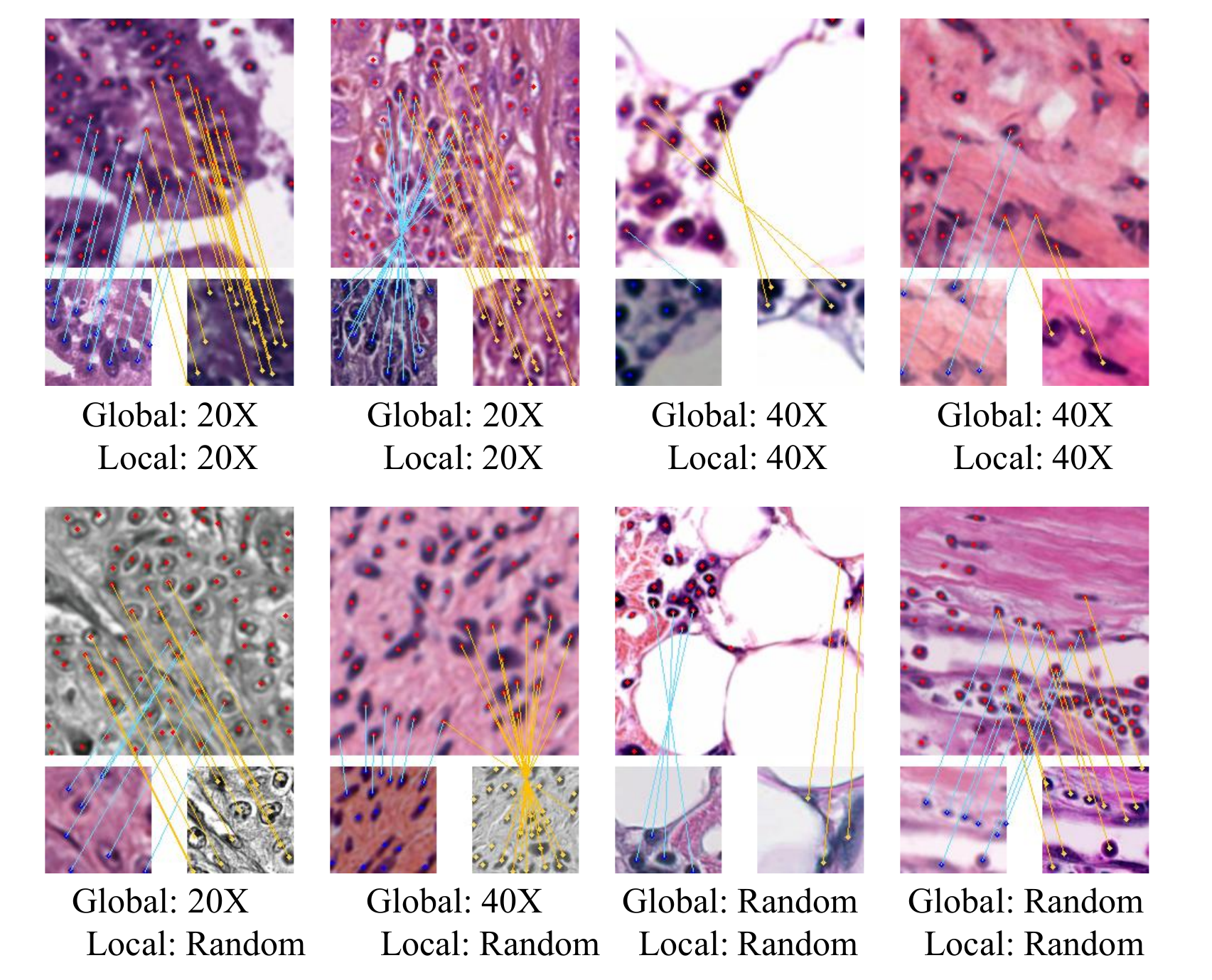}
    \caption{
    \textbf{MPP-Based Cropping.}
    Samples at specific physical scales are precisely cropped using MPP for MUSE. Each sample presents one global view and two local views.
    }
    \label{fig:multi-crop}
\end{figure}

\noindent \textbf{MPP-Based Cropping.}
Pretraining samples for MUSE are illustrated in Figure \ref{fig:multi-crop}. 
Based on nucleus coordinates, we accurately match local regions between paired samples after spatial transformations. 
This approach enables the use of complex spatial transformations and scale changes in local self-distillation, thereby enhancing the local representation of models.

\begin{figure*}[t]
    \centering
    \includegraphics[width=0.99\linewidth]{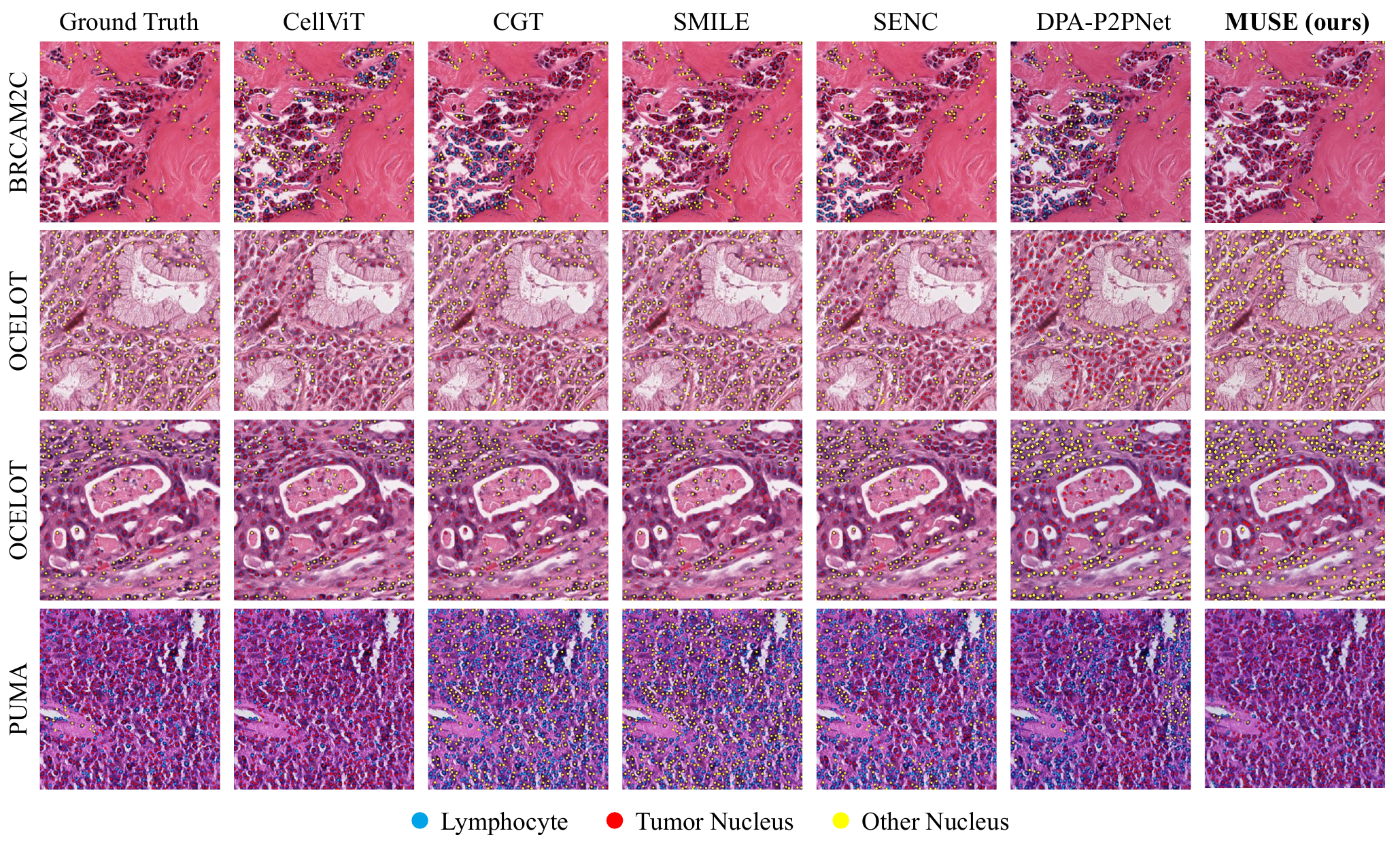}
    \caption{
    \textbf{Visualization of nucleus detection and classification results.}
    MUSE shows strong nucleus detection and classification performance across all three benchmark datasets.
    }
    \label{fig:ndc_vis}
\end{figure*}

\noindent \textbf{Nucleus Detection and Classification (NDC).}
Figure \ref{fig:ndc_vis} visualizes the results of methods in NDC. 
Other methods exhibit numerous classification errors, whereas MUSE accurately distinguishes nucleus types.

\section{Appendix B: Detailed Experiment Settings}


\subsection{Dataset}
$\text{TCGA}_{\text{nu}}$ is constructed from 11 types of cancer in TCGA~\cite{liu2018integrated} through a four-step process.
First, nucleus detection is performed for each WSI with a ResNet-18 trained on BRCAM2C.
Second, 2048-pixel ROIs at 40x magnification are patched from WSIs and further form a candidate sample set.
Third, after excluding ROIs containing fewer than 20 nuclei, random sampling is employed to obtain a balanced dataset across cancer types, with a total of 500K samples.
Finally, to ensure no data leakage occurs in pertaining, all ROIs from the same WSI as any sample in BRCAM2C~\cite{mcspatnet} or OCELOT~\cite{ryu2023ocelot} are filtered out.
No additional filtering based on PUMA~\cite{puma}, which is constructed with non-TCGA WSIs.
The resulting $\text{TCGA}_{\text{nu}}$ dataset contains 483,627 samples.
The full names, cancer abbreviations, and sample counts for each cancer type included in $\text{TCGA}_{\text{nu}}$ are listed in Table \ref{tb:tcga_nu}.

\subsection{MUSE}

\noindent \textbf{Pretraining.}
For MUSE, we mainly follow the DINO~\cite{dino} hyperparameter settings for ResNet-50, ViT-S, and ViT-B.
The main differences are: 1) the batch size is set to 256, 2) the total number of iterations is 132K, 3) the number of warm-up steps is 9.4K, and 4) the teacher temperature starts at 0.04 and ends at 0.05.
In addition, the learning rates for ResNet-50 and ViT are set to $5 \times 10^{-3}$ and $1 \times 10^{-3}$, respectively. 
All ablation experiments follow the same hyperparameter settings. 
For LFoV-MUSE, we resume the MUSE pretrained model and further pretrain it for an additional 29K iterations.

\noindent \textbf{Fine-Tuning.}
Baselines are implemented with their released codes and the default hyperparameters.
Finetuning MUSE on BRCAM2C is optimized using Adam~\cite{liu2019variance}, a batch size of 4, a learning rate of $1 \times 10^{-4}$, cosine annealing learning rate decay to $1 \times 10^{-5}$, and epochs of 150.
As OCELOT and PUMA include more samples, finetuning MUSE on these two datasets is optimized using Adam, a batch size of 4,  a learning rate of $1 \times 10^{-4}$, cosine annealing learning rate decay to $1 \times 10^{-6}$, and epochs of 100.
For $\mathcal{L}_{cons}$, the loss weight $\lambda_{cons}$ is gradually increased from 0 to 0.1 based on $1-\cos(\pi \cdot i / N_{epoch})$, where $i$ and $N_{epoch}$ denote the current epoch and max epoch, respectively.
L2 loss and cross-entropy loss are employed to implement $\lambda_{reg}$ and $\lambda_{type}$, respectively.
Furthermore, the weights for the coordinate regression loss and the classification loss are set to $5\times10^{-3}$ and $1.0$, respectively.

\noindent \textbf{Computing Infrastructure.}
All experiments are conducted on NVIDIA H20 GPUs. For each GPU, 24 CPU cores and 230 GB of memory are allocated. Specifically, pretraining utilizes 16 NVIDIA H20 GPUs. Pretraining based on ResNet-50, ViT-S, and ViT-B requires 544, 320, and 480 GPU hours, respectively. LFoV-MUSE based on ResNet-50, ViT-S, and ViT-B consumes 272, 192, and 320 GPU hours, respectively. Finetuning is performed on a single NVIDIA H20 GPU.
All implementations are based on Torch 2.2.2.

\subsection{Evaluation}

\noindent \textbf{Dense Prediction.}
For nucleus classification, evaluations are performed in three steps: 1) obtaining the feature map, 2) extracting feature vectors with nucleus coordinates from the feature map via interpolation, and 3) using these feature vectors for evaluation.
The first step is adapted according to the pretrained model architecture to obtain an optimal feature map. 
For ViT, the final output token sequence is reassembled into a feature map based on the patch size. 
For hierarchical architectures such as ResNet and Swin Transformer~\cite{liu2021swin}, feature maps are extracted from each block, interpolated to a common size, and then concatenated to form the final feature map.
In addition, the original inference procedure of each baseline is used for nucleus detection and classification.







    
    

    

\begin{table}[!t]
    \centering
    \small
    \renewcommand{\tabcolsep}{9pt}
    \renewcommand{\arraystretch}{2}

    \resizebox{\linewidth}{!}{

    \begin{tabular}{ccc}

    \toprule

    Full Name & Abbreviation & $N_s$ \\

    \midrule

    Bladder Urothelial Carcinoma & BLCA &  40167\\
    Breast Invasive Carcinoma & BRCA &  43437\\
    Colon Adenocarcinoma & COAD &  45454\\
    Head and Neck Squamous Cell Carcinoma & HNSC &  43900\\
    Kidney Renal Clear Cell Carcinoma & KIRC & 44068\\
    Lung Adenocarcinoma & LUAD &  45454\\
    Lung Squamous Cell Carcinoma & LUSC &  45454\\
    Pancreatic Adenocarcinoma & PAAD &  45454\\
    Rectum Adenocarcinoma & READ &  45454\\
    Stomach Adenocarcinoma & STAD &  43728\\
    Uterine Corpus Endometrial Carcinoma & UCEC &  41057\\
    - & Total & 483627 \\

    \bottomrule
    
    \end{tabular}
    }

    \caption{
    \textbf{Sample statistics.} The full name, cancer abbreviation, and total sample count of each cancer in $\text{TCGA}_{\text{nu}}$.
    }
    \label{tb:tcga_nu}
    
\end{table}

\noindent \textbf{KNN.}
For each KNN evaluation, we evaluate with k = 10, 20, 100, 200, and 500, and report the best value.

\noindent \textbf{Linear Probing.}
For each linear probing evaluation, the backbone is frozen, and the linear classifier weights are initialized from $\mathcal{N}(0, 0.01)$ while the bias is initialized to 0. Optimization is performed using SGD with a learning rate annealed from 0.01 to 0 via cosine annealing. The number of epochs is set to 100 and the batch size to 256.

\noindent \textbf{Fine-Tuning.}
We initialize the linear classifier with the parameters obtained from linear probing to facilitate more effective fine-tuning of the backbone. 
AdamW is used as the optimizer with a learning rate of $1\times10^{-5}$, for a total of 10 epochs and a batch size of 32.

\noindent \textbf{F1 Score.}
We follow the common practice~\cite{mcspatnet,dpap2pnet} in nucleus detection and classification by determining one-to-one matches between predictions and ground truth based on distance. After matching, the F1 score for each nucleus type is calculated based on the number of predictions, the number of ground truth nuclei, and the number of correct predictions. The average F1 score is then obtained by averaging the F1 scores across all classes.

\newpage

\newpage

\bibliography{aaai2026}

\end{document}